\documentclass[10pt,twocolumn,letterpaper]{article}

\usepackage{iccv}
\usepackage{times}
\usepackage{epsfig}
\usepackage{graphicx}
\usepackage{amsmath}
\usepackage{amssymb}

\usepackage{blindtext}
\usepackage{caption}
\usepackage{amsfonts}
\usepackage[inline]{enumitem}
\usepackage{algorithmic}
\usepackage{textcomp}

\usepackage{booktabs}
\usepackage{multirow}
\usepackage{float}
\usepackage{adjustbox}
\usepackage{etoolbox}
\usepackage{dsfont}

\usepackage{float}

\usepackage{lipsum}
\usepackage{stfloats}
\usepackage{multicol}
\usepackage{amsmath,bm}

\usepackage{array}

\usepackage{tabulary}
\usepackage{xcolor}

\usepackage{paralist}

\usepackage[pagebackref=true,breaklinks=true,letterpaper=true,colorlinks,bookmarks=false]{hyperref}

\iccvfinalcopy 

\usepackage[pagebackref=true,breaklinks=true,colorlinks,bookmarks=false]{hyperref}
\hypersetup{colorlinks, citecolor=blue}
\hypersetup{colorlinks, linkcolor=blue}

\ificcvfinal\fi

\begin{document}

\title{HIDA: Towards Holistic Indoor Understanding for the Visually Impaired via Semantic Instance Segmentation with a Wearable Solid-State LiDAR Sensor}

\author{Huayao Liu, Ruiping Liu, Kailun Yang\thanks{Corresponding author (e-mail: {\tt kailun.yang@kit.edu}).}, Jiaming Zhang, Kunyu Peng and Rainer Stiefelhagen\\
Karlsruhe Institute of Technology
}

\maketitle
\ificcvfinal\thispagestyle{empty}\fi

\begin{abstract} 
Independently exploring unknown spaces or finding objects in an indoor environment is a daily but challenging task for visually impaired people. However, common 2D assistive systems lack depth relationships between various objects, resulting in difficulty to obtain accurate spatial layout and relative positions of objects. To tackle these issues, we propose HIDA, a lightweight assistive system based on 3D point cloud instance segmentation with a solid-state LiDAR sensor, for holistic indoor detection and avoidance. Our entire system consists of three hardware components, two interactive functions~(obstacle avoidance and object finding) and a voice user interface. Based on voice guidance, the point cloud from the most recent state of the changing indoor environment is captured through an on-site scanning performed by the user. In addition, we design a point cloud segmentation model with dual lightweight decoders for semantic and offset predictions, which satisfies the efficiency of the whole system. After the 3D instance segmentation, we post-process the segmented point cloud by removing outliers and projecting all points onto a top-view 2D map representation. The system integrates the information above and interacts with users intuitively by acoustic feedback. The proposed 3D instance segmentation model has achieved state-of-the-art performance on ScanNet v2 dataset. Comprehensive field tests with various tasks in a user study verify the usability and effectiveness of our system for assisting visually impaired people in holistic indoor understanding, obstacle avoidance and object search. 
\end{abstract}

\section{Introduction}

For sighted people, when they enter an unfamiliar indoor environment, they can observe and perceive the surrounding environments through their vision. However, such a global scene understanding is challenging for people with visual impairments. They often need to approach and touch the objects in the room one by one to distinguish their categories and get familiar with their locations. This is not only inconvenient but also creates some risks for visually impaired people in their everyday living and travelling tasks. In this work, we develop a system to help vision-impaired people understand unfamiliar indoor scenes.

Some assistance systems leverage various sensors (such as radar, ultrasonic, and range sensors) to help the vision-impaired avoid obstacles~\cite{bai2017smart,katzschmann2018safe,long2019unifying,zhang2017indoor}. With the development of deep learning, vision tasks like object detection and image segmentation can yield precise scene perception. Different vision-based systems were proposed towards environment perception and navigation assistance for visually impaired people. However, most 2D image semantic-segmentation-based systems~\cite{lin2019deep,wang2018environmental,yang2018unifying} and 3D-vision-based systems~\cite{bauer2020enhancing,caraiman2017computer,yang2015new} cannot provide a holistic understanding, because these systems only process the current image or image with depth information captured by the camera, rather than a complete scan of the surroundings.

\begin{figure*}[t]
\begin{center}
   \includegraphics[width=1\linewidth]{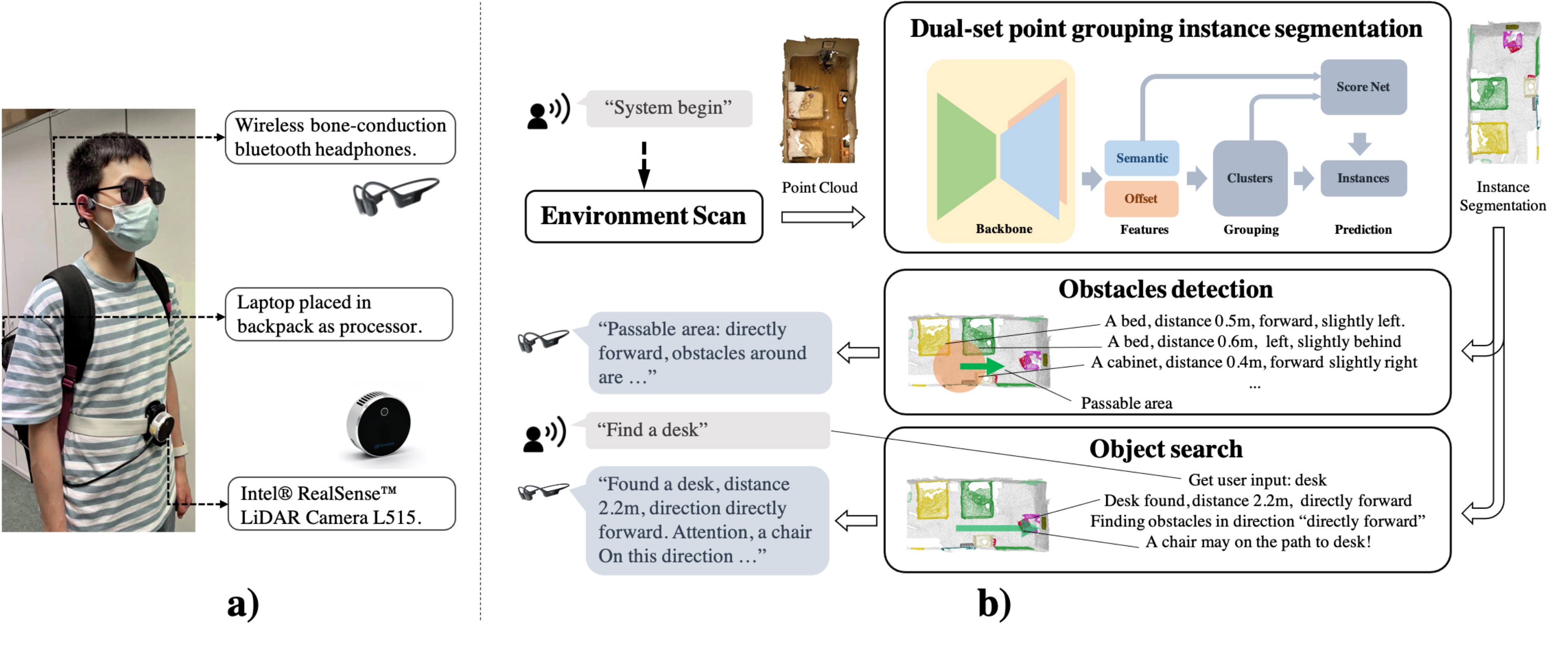}
\end{center}
\vskip-5ex
   \caption{\textbf{a)} Hardware overview. \textbf{b)} System overview.}
\label{overview}
\vskip-3ex
\end{figure*}

Compared with 2D images, 3D point clouds contain more information and are suitable for reconstruction of the surrounding environment. Thereby, in this work, we propose HIDA, an assistance system for Holistic Indoor Detection and Avoidance based on semantic instance segmentation. The main structure of HIDA is shown in Fig.~\ref{overview}. When the user enters an unfamiliar room, the user can wake up the system by voice, and then the system will help the user to scan the surroundings by running Simultaneous Localization and Mapping (SLAM). Then the obtained point cloud will be delivered into an instance segmentation network, where each point will be attached with instance information. We adapt PointGroup~\cite{pointgroup} in our instance segmentation network and modify the structure to acquire 3D semantic perception with higher precision. We enable a cross-dimension understanding by converting the point cloud with instance-aware semantics to a usable and suitable representation, \ie, 2D top-view segmentation, for assistance. After reading the current user location, the system will inform vision-impaired people through voice about the obstacles around and suggest the safe passable direction. In addition, the user can specify a certain type of object in the room, and our system will tell the user the distance and direction of this object and warn possible obstacles on the path.

Our segmentation model is trained and evaluated on the ScanNet v2 dataset~\cite{scannet}, which yields an accurate and robust surrounding perception. During field experiment and various pre-tests, the learned model performed satisfactorily in our indoor usage scenarios, which makes it suitable for real-world applications. Even when an object was partially scanned, this network is still able to recognize and render relatively accurate classification. To evaluate the assistance functions of our system, we designed different tasks. Our system has achieved significant results in reducing collisions with obstacles. According to the questionnaire survey, the users believe that our system will help vision-impaired people in indoor scene understanding. To the best of our knowledge, we are the first to use 3D semantic instance segmentation for assisting the visually impaired.

In summary, we deliver the following contributions:
\begin{compactitem}
    \item We propose HIDA, a wearable system with a solid-state LiDAR sensor, which helps the visually impaired to obtain a holistic understanding of indoor surroundings with object detection and obstacle avoidance. 
    \item We designed a 3D instance segmentation model, achieving the state-of-the-art in mAP on ScanNet v2.
    \item We convert the point cloud with semantic instance information to a usable top-view representation suitable for assisting vision-impaired people.
    \item We conducted user studies to evaluate obstacle avoidance and object search, verifying the usability and benefit of the proposed assistance system.
\end{compactitem}

\section{Related Work}

\noindent\textbf{Semantic Segmentation for Visual Assistance.}
Since the surge of deep learning particularly the concept of fully convolutional networks~\cite{fcn}, semantic segmentation can be performed end-to-end, which enables a dense surrounding understanding. Thereby, semantic segmentation has been introduced into vision-based navigational perception and assistance systems~\cite{duh2020v,hsieh2020development,lin2019deep,mahendran2021computer,miksik2015semantic,watson2020footprints,zhang2021perception}.

Yang~\etal~\cite{yang2018unifying} put forth a real-time semantic segmentation architecture to enable universal terrain perception, which has been integrated in a pair of stereo-camera glasses and coupled with depth-based close obstacle segmentation. Mao~\etal~\cite{mao2021panoptic} designed a panoptic lintention network to reduce the computation complexity in panoptic segmentation for efficient navigational perception, which unifies semantic and instance-specific understanding. Semantic segmentation has also been leveraged to address intersection perception with lightweight network designs~\cite{cao2020rapid,hsieh2020outdoor}, whereas most instance-segmentation-based assistance systems~\cite{long2019unifying,yohannes2019content} directly use the classic Mask R-CNN model~\cite{he2017mask} and rely on sensor fusion to output distance information.

Compared to 2D segmentation-driven assistance, 3D scene parsing systems~\cite{caraiman2017computer,yang2015new,zatout2019ego} fall behind, as these classical point cloud segmentation works focus on designing principled algorithms for walkable area detection~\cite{aladren2014navigation,wang2017enabling,zatout2019ego} or stairs navigation~\cite{perez2017stairs,yang2015new,ye20173}. In this work, we devise a 3D semantic instance segmentation system for helping visually impaired people perceive the entire surrounding and provide a top-view understanding, which is critical for various indoor travelling and mapping tasks~\cite{hu2019indoor,li2016isana,li2018semantic,liu2020indoor}. 

\noindent\textbf{3D Semantic and Instance Segmentation.}
With the appearance of large-scale indoor 3D segmentation datasets~\cite{S3DIS,scannet}, point cloud semantic instance segmentation becomes increasingly popular. It allows to go beyond 2D segmentation and render both point-wise and instance-aware understanding, which is appealing for assisting the visually impaired. Early works follow two mainstreams. One is based on object detection that first extracts 3D bounding boxes and then predicts point-level masks~\cite{hou20193d,yang2019learning,yi2019gspn}. Another prominent paradigm is segmentation-driven, which first infers semantic labels and then groups points into instances by using point embedding representations~\cite{lahoud20193d,liu2019masc,pham2019jsis3d,wang2018sgpn,wang2019associatively}.

Recently, PointGroup~\cite{pointgroup} is designed to enable better grouping of points into semantic objects and separation of adjacent instances. DyCo3D~\cite{dyco3d} employs dynamic convolution customized for 3D instance segmentation. OccuSeg~\cite{han2020occuseg} relies on multi-task learning by coupling embedding learning and occupancy regression. 3D-MPA~\cite{engelmann20203d} generates instance proposals in an object-centric manner, followed by a graph convolutional model enabling higher-level interactions between nearby instances. Additional methods use panoptic fusion~\cite{jaus2021panoramic,narita2019panopticfusion,wu2021scenegraphfusion}, transformers~\cite{guo2020pct,zhao2020point} and omni-supervision~\cite{omni_point,yang2020omnisupervised} towards complete understanding.

In this work, we build a holistic semantic instance-aware scene parsing system to help visually impaired people understand the entire surrounding. We augment instance segmentation with a lightweight dual-decoder design to better predict semantic- and offset features. Differing from other cross-dimension~\cite{hu2021bidirectional,liu20213d} and cross-view~\cite{cartillier2020semantic,pan2020cross,peng2021mass} sensing platforms, we aim for top-view understanding and our holistic instance-aware solution directly leverages 3D point cloud segmentation results and aligns them onto 2D top-view representations for generating assistive feedback.

\section{HIDA: Proposed System}

The entire architecture of HIDA is depicted in Fig.~\ref{overview}, including hardware components, user interfaces, and the algorithm pipeline. Designed to maximize the stability of the portable system, only very few parts are integrated into our prototype. As shown in Fig.~\ref{overview}a), the system is composed of three hardware components. First, a lightweight solid-state LiDAR sensor is attached to a belt for collecting point clouds. The RealSense L515, as the world's smallest high-resolution LiDAR depth camera, is well suitable as part of wearable devices. In addition, the scanning range up to $9m$ is suitable for most indoor scenes. A laptop placed in a backpack is the second component of our system. The laptop with a GPU processor ensures that the instance segmentation can be performed in an online manner. As for input and output interfaces, a bone-conduction headset with a microphone is the third component. The audio commands from users can be recognized by the user interface. Also beneficially, thanks to the bone-conduction earphones, internal acoustic cues from our system and external environmental sounds can be separately perceived by the users, which is safety-critical for assisting the visually impaired.

\begin{figure}[t]
\begin{center}
    \scalebox{0.9}[0.8]{
        \includegraphics[width=1\linewidth]{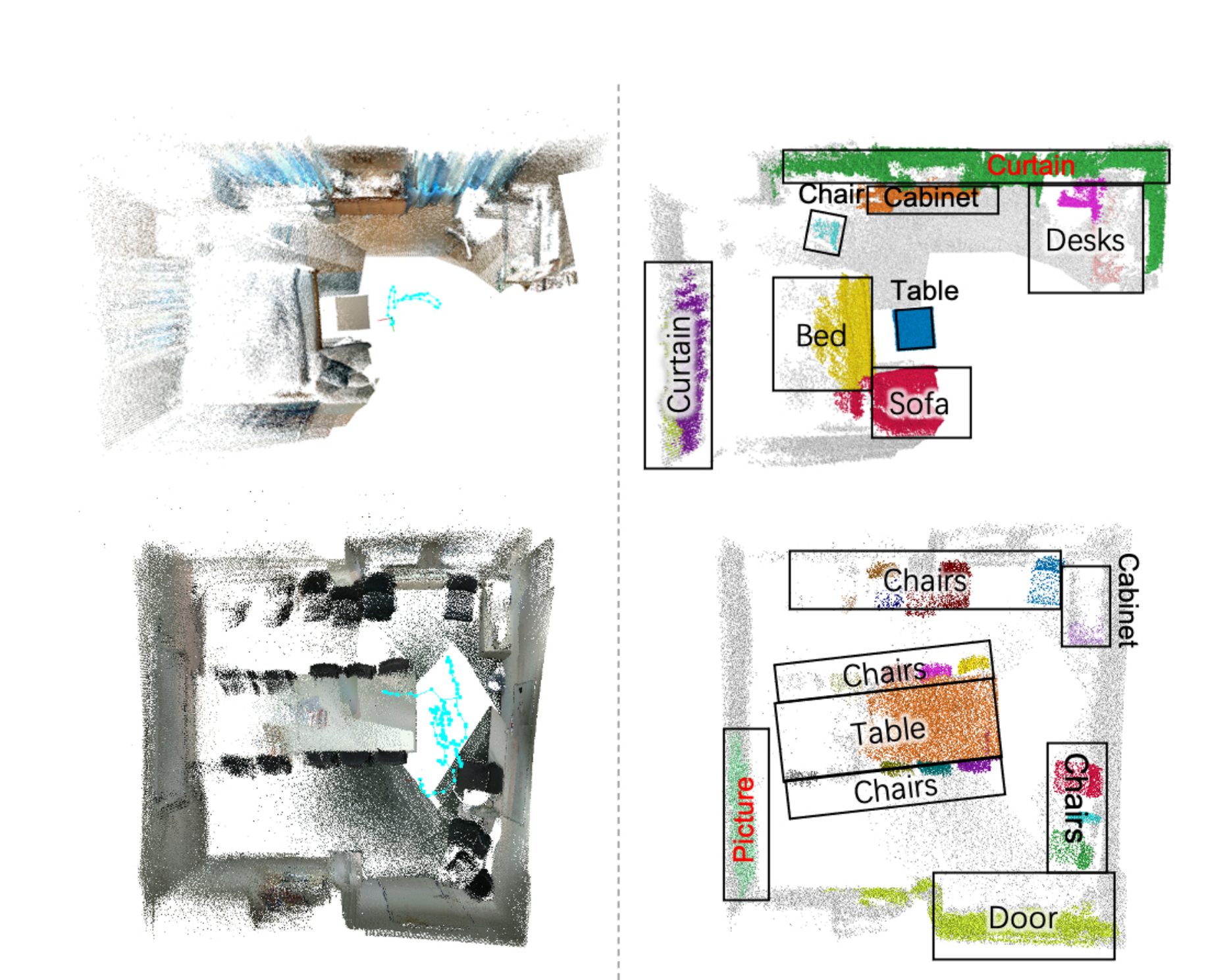}
    }
\end{center}
\vskip-3ex
   \caption{\textbf{Left:} Capturing point cloud within a limited moving range. \textbf{Right:} Instance segmentation results (red colored indices wrongly classified instances).}
\label{slam}
\vskip-4ex
\end{figure}

\subsection{Environment Scan}
The users will collect independently the point cloud under the audio guidance. At the same time, the system also needs to obtain the user's position in the point cloud map. These can be achieved through Simultaneous Localization and Mapping (SLAM) with odometry~\cite{chen2021panoramic,labbe2019rtab,wang2021lightweight}. There are many mature SLAM frameworks. However, for the visually impaired, collecting point clouds in an unfamiliar environment is different from conventional ways. Specifically, the entire room is usually scanned at the entrance, so the user's movement is limited in a small range and those movement will mostly be the in-situ rotation. In addition, the motion of the human body is more unstable compared to robots. In this case, the odometry of SLAM may lose tracking. If this happens, the camera is required to move back to the previous position of the keyframe for loop closing, which is hard for vision-impaired people. Therefore, we value more the robustness of the mapping process. In our field test, RTAB-Map~\cite{labbe2019rtab} achieved a reliable performance under such requirements. Fig.~\ref{slam} shows two scanning results in real-world scenes using RTAB-Map. It can be seen that even in a limited range of movement, HIDA can obtain point clouds with rich and dense object information. After proficiency, users can complete the collection of dense point clouds in most cases by their own, even if they cannot check the collection of point clouds through the display at the same time.

In our system, once the ``start'' instruction signal is recognized by the voice user interface, the environment scan process will begin. The odometry will update the current position and direction of the user at the same time. Under the audio guidance, the user can slowly turn around to scan the surrounding environment. In practice, the default scan time of a whole room is set as $20$ seconds, which can obtain a sufficient amount of keyframes from different directions. After that, the captured point cloud will be sent to the segmentation network for 3D instance segmentation.

\subsection{3D Instance Segmentation}

\begin{figure*}[t]
\begin{center}
    \scalebox{0.9}[0.9]{
        \includegraphics[width=1\linewidth]{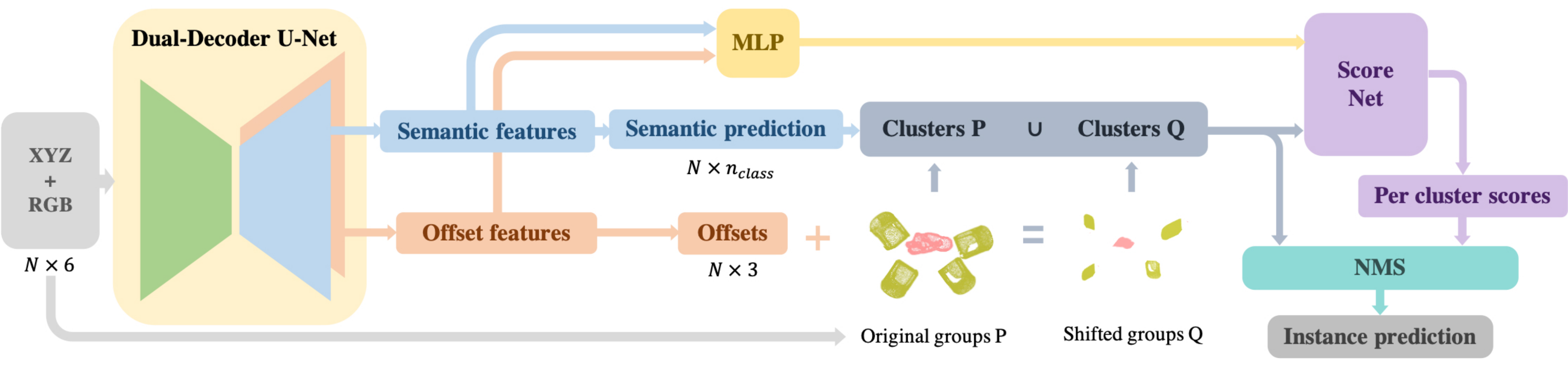}
    }
\end{center}
   \vskip-4ex
   \caption{Architecture of our 3D point cloud segmentation model with dual-decoder U-Net in the backbone structure to separately predict semantic- and offset features, and subsequent adaptations for holistic instance-aware semantic understanding.
   }
\label{model}
\vskip-3ex
\end{figure*}

\begin{figure}[t]
\begin{center}
   \includegraphics[width=1\linewidth]{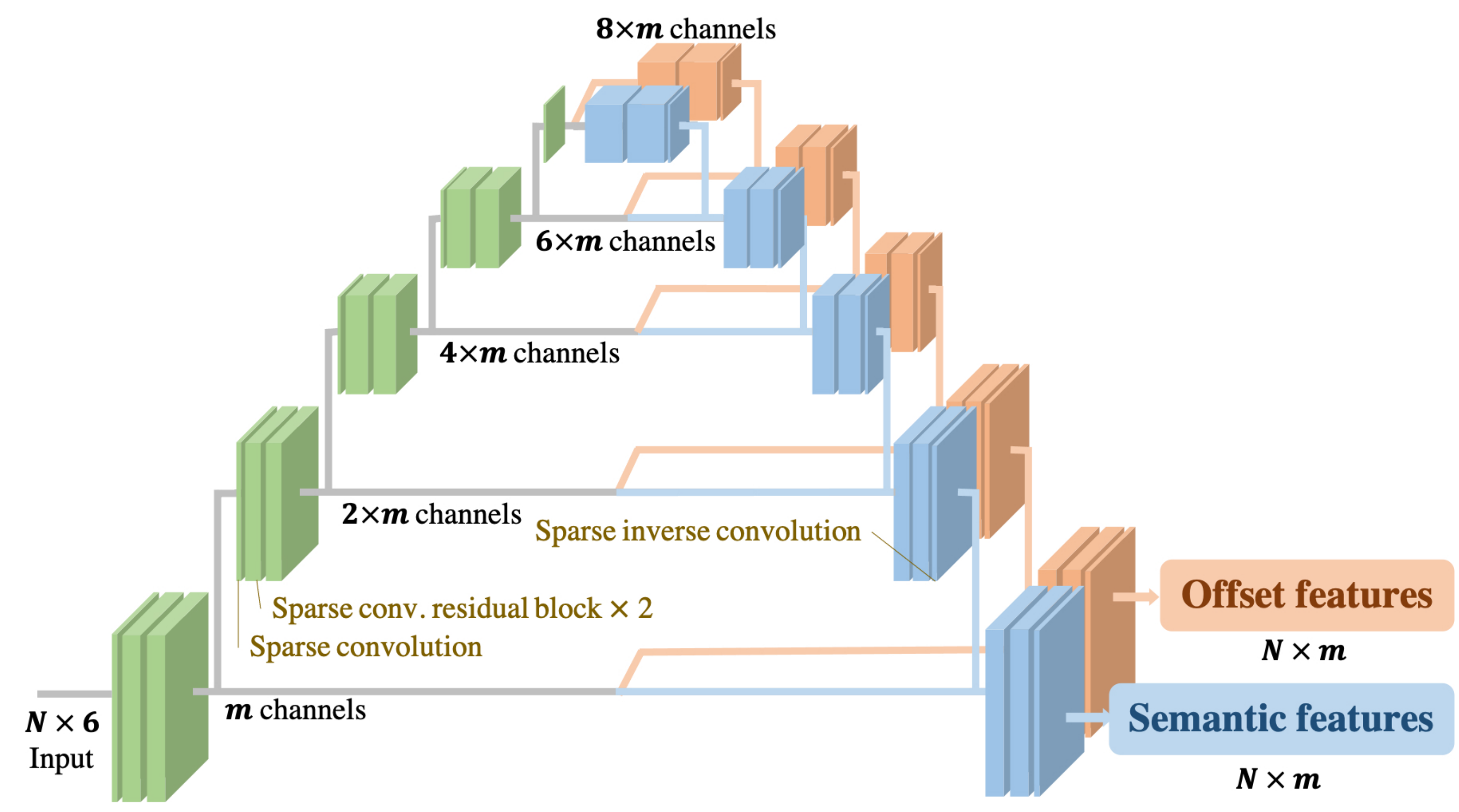}
\end{center}
   \vskip-4ex
   \caption{Backbone architecture. $X \times m$ indicates the number of encoder output channels. We set $m=16$ or $m=32$.}
\label{ddunet}
\vskip-3ex
\end{figure}

In the data preprocessing stage, if the point cloud is too large, the segmentation time and memory requirement  will be significantly increased. Considering the efficiency of 3D segmentation, we restrict the amount of points under $200,000$. If the total number of points in the point cloud exceeds $200,000$, we will downsample them evenly for saving running time and memory usage. Besides, some noise points will be removed as outliers which is caused by the influence of natural lighting.

Inspired by PointGroup~\cite{pointgroup}, we design a 3D instance segmentation architecture. PointGroup uses a $7$-layer sparse convolution U-Net~\cite{ronneberger2015u} to extract features. Each layer consists of an encoder and a decoder. The extracted feature will be decoded into $2$ branches: semantic branch and offset branch. The semantic branch builds the clusters that have the same semantic labels. The offset branch predicts per-point offset vectors to shift each point towards the instance centroid and builds shifted point clusters that belong to the same instances. Those cluster proposals will be voxelized and then scored. In the inference, Non-Maximum Suppression~(NMS) is performed to make final instance predictions.

Different from~\cite{pointgroup}, our key adaptation lies in the backbone structure in order to enable a clear separation of dense instances and reach a better performance on semantic and offset predictions, which are essential in 3D instance segmentation. The architecture of our modified model is shown in Fig.~\ref{model}. Specifically, we use one encoder and dual decoders in each layer, including a decoder for semantic features and another for offset features. They are extracted separately in the early stage of the network. In addition, this dual-decoder U-Net consists of only $5$ layers instead of $7$, which can reduce parameters and speed up the inference. More details are shown in Fig.~\ref{ddunet}. This architecture has a clearly better performance compared with the original backbone, while maintaining a low amount of parameters.

As mentioned before, many objects may only be partially scanned. Even so, the model can still accurately identify most of the objects in the point cloud. We show some examples of instance segmentation results in Fig.~\ref{slam}.

\subsection{User Interface}

\begin{figure}[t]
\vskip-2ex
\begin{center}
        \includegraphics[width=1\linewidth]{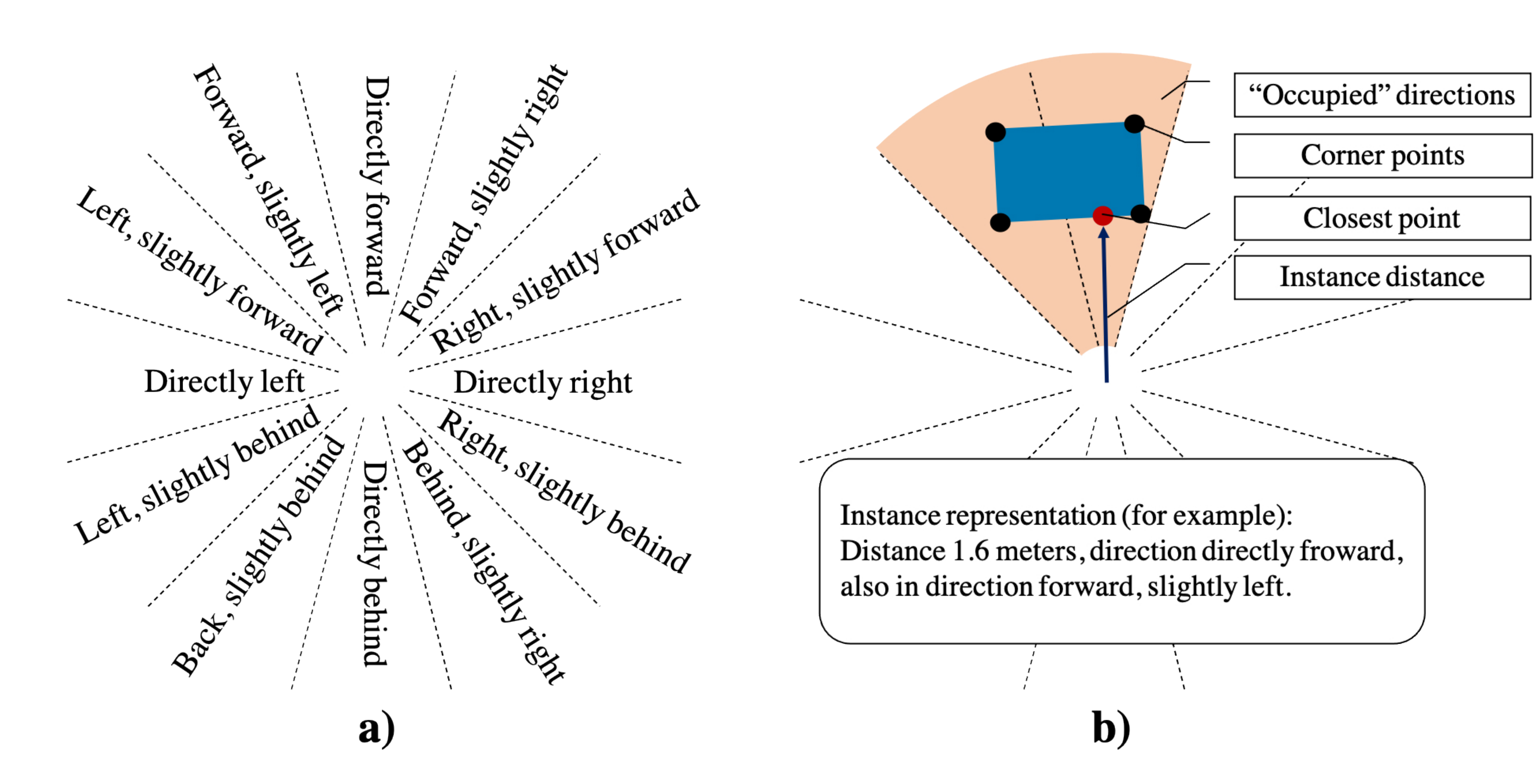}
\end{center}
   \vskip-5ex
   \caption{Representation of instance distance and direction.}
\label{represent}
\vskip-3ex
\end{figure}

After capturing point cloud and obtaining semantic and instance information, we can actually provide a variety of information to vision-impaired people. Although we have already downsampled the point cloud before, the calculation with the whole point cloud is still very time-consuming. In order to reduce the processing time, the proposed system only traverses the entire point cloud once to extract information of interest of the users. The proposed implementation is as follows: First, all points will be projected onto a plane parallel to the ground ($XY$-plane). Then, the current user location will be updated. For each instance, $5$ feature points will be searched in points belonging to the same segmented instance: one point closest to the user and four coordinate extreme points (corner points) of this object. The direction will be calculated according to the camera pose information. We build the 2D camera coordinate $X^{'}Y^{'}$ relative to the $XY$-plane, and then transform the feature points coordinate $(x_{i}, y_{i})$ into $(x^{'}_{i}, y^{'}_{i})$. $+x^{'}$ corresponds to the front of the user, and $+y^{'}$ corresponds to the left. The coordinates are not intuitive for the user, so our designed user interface will represent instances with \emph{distance} and \emph{direction} relative to the user. The directions will be divided into $12$ areas, as shown in Fig.~\ref{represent}a). In this way, the location of each instance will be replaced by the distance and direction of the point on the instance that is the closest to the user. Besides, in order to help the user bypass the obstacle, the direction of the ``corner points'' will also be denoted as the direction ``occupied'' by the object. An example is shown in Fig.~\ref{represent}b).

\begin{figure}[t]
\begin{center}
   \includegraphics[width=0.8\linewidth]{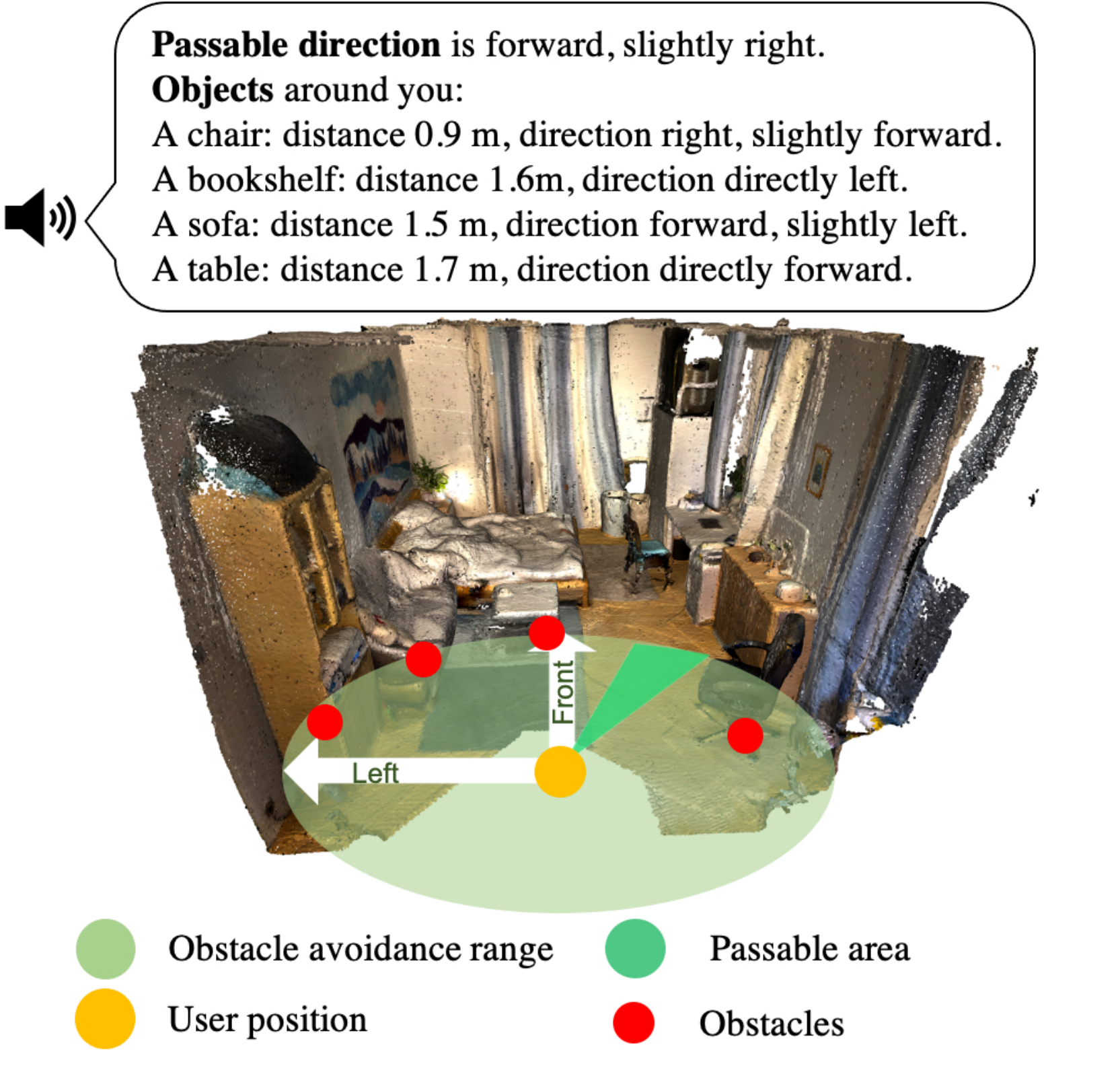}
\end{center}
   \vskip-4ex
   \caption{Obstacle avoidance in an example scene.}
\label{mode1}
\vskip-4ex
\end{figure}

\noindent\textbf{Obstacle avoidance.}
Furthermore, we proposed two interactive functions. The first function is the obstacle avoidance. The user will set an obstacle avoidance range. The system will primarily eliminate the direction occupied by obstacles within this range to find passable area. If all directions in the scanned area are already occupied by the obstacles, not scanned area will be directed and suggested to users as potential passable area. The passable area and all objects information within the detection range will be output. Fig.~\ref{mode1} is a functional example in an indoor scene. 

\begin{figure}[t]
\begin{center}
    \scalebox{0.96}[0.96]{
        \includegraphics[width=0.8\linewidth]{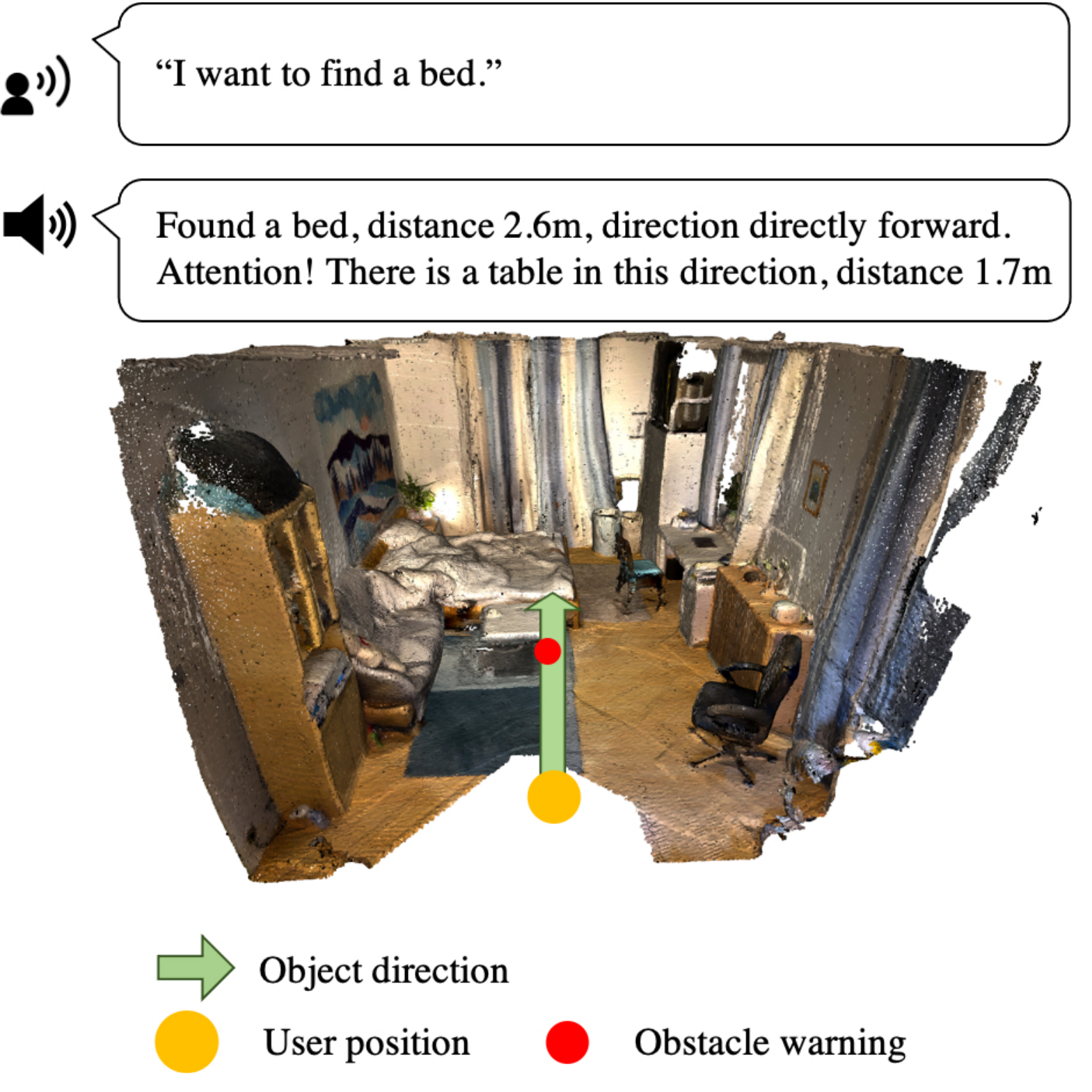}
    }
\end{center}
   \vskip-4ex
   \caption{Object search in an example scene.}
\label{mode2}
\vskip-4ex
\end{figure}

\noindent\textbf{Object finding.}
The second function is the object search. The user specifies an object category of interest through voice commands, such as a ``Find a desk'' instruction. Then, the system will search for the corresponding instance and return the object position through acoustic cues. For instance, ``Found a desk, distance $2.2$ meters, direction in directly forward'' will be output as speech via the bone-conducted headset. In addition, in order to help users navigate to the object, our system can also alert obstacles in the direction leading to the object. For example, ``Attention, a chair in this direction, distance $1.3$ meters''. The detailed schematic view is in Fig.~\ref{mode2}.

\section{Evaluation}

\subsection{Experiments}

\subsubsection{Quantitative results of segmentation model}

\begin{figure}[h]
\vskip-2ex
\begin{center}
    \includegraphics[width=1\linewidth]{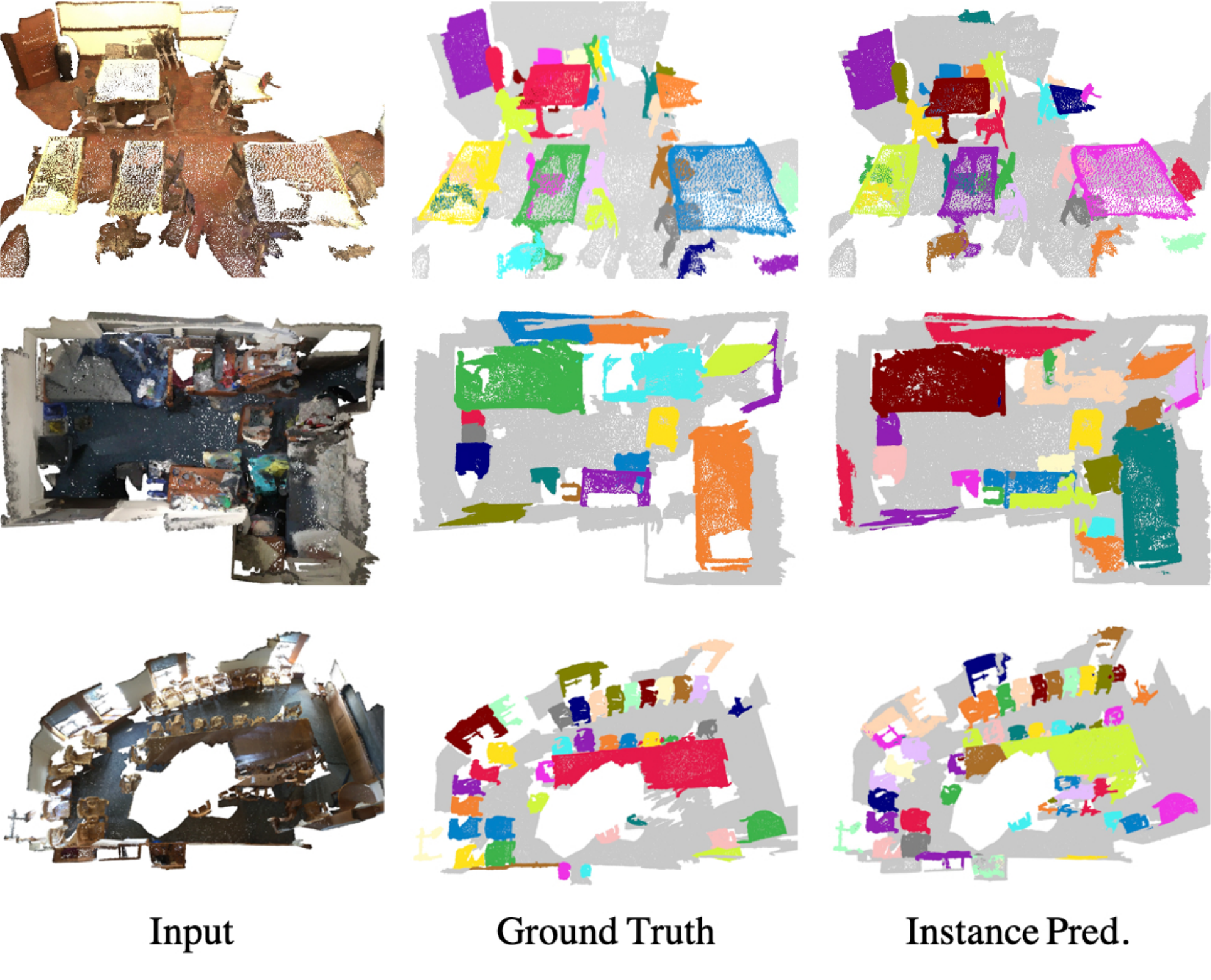}
    \end{center}
    \vskip-4ex
    \caption{Qualitative results on ScanNetV2 validation set.}
\label{scannet}
\vskip-3ex
\end{figure}

We trained our instance segmentation model on ScanNet v2 dataset~\cite{scannet}, containing $1613$ scans of indoor scenes. The dataset is spitted into $1201$, $312$, and $100$ scans in the training, validation, and testing subsets, respectively. We set the cluster voxel size as $0.02m$, and cluster radius as $0.03m$ and the minimum cluster point number as $50$. In the training process, we uses Adam optimizer~\cite{adam_optimization} with a learning rate of $0.001$. Our model learned through the entire training set for $360$ times with a batch size of $8$. We report the quantitative performance in Table~\ref{tab:scannet_test} and Table~\ref{tab:scannet_val}. Several qualitative visualizations of the segmentation results are shown in Fig.~\ref{scannet}. In these three scenes, the instances are placed quite densely (\eg, many chairs are placed side by side in the third scenes), and some instances have not been completely scanned. But our model can still separate them well, which is beneficial for upper-level assistance.

\textbf{\begin{table*}[t]
        \resizebox{\textwidth}{!}{
                \begin{tabular}{ r |ccc|cccccccccccccccccc}
                        \toprule
                        & \textbf{mAP} & \textbf{$AP_{50}$} & \textbf{$AP_{25}$}& \rotatebox{90}{bathhub} & \rotatebox{90}{bed} & \rotatebox{90}{bookshelf} & \rotatebox{90}{cabinet} & \rotatebox{90}{chair} & \rotatebox{90}{counter} & \rotatebox{90}{curtain} & \rotatebox{90}{desk} & \rotatebox{90}{door} & \rotatebox{90}{otherfu.} & \rotatebox{90}{picture} & \rotatebox{90}{refrigerator} & \rotatebox{90}{s.curtain} & \rotatebox{90}{sink} & \rotatebox{90}{sofa} & \rotatebox{90}{table} & \rotatebox{90}{toilet} & \rotatebox{90}{window} \\
                        \midrule
                        3D-SIS~\cite{hou20193d} & 16.1 & 38.2 & 55.8 & 40.7 & 15.5 & 6.8 & 4.3 & 34.6 & 0.1 & 13.4 & 0.5 & 8.8 & 10.6 & 3.7 & 13.5 & 32.1 & 2.8 & 33.9 & 11.6 & 46.6 & 9.3 \\
                        3D-BoNet~\cite{yang2019learning} & 25.3 & 48.8 & 68.7 & 51.9 & 32.4 & 25.1 & 13.7 & 34.5 & 3.1 & 41.9 & 6.9 & 16.2 & 13.1 & 5.2 & 20.2 & 33.8 & 14.7 & 30.1 & 30.3 & 65.1 & 17.8 \\
                        MASC~\cite{liu2019masc} & 25.4 & 44.7 & 61.5 & 46.3 & 24.9 & 11.3 & 16.7 & 41.2 & 0.0 & 34.5 & 7.3 & 17.3 & 24.3 & 13.0 & 22.8 & 36.8 & 16.0 & 35.6 & 20.8 & 71.1 & 13.6 \\
                        SALoss-ResNet~\cite{liang20203d} & 26.2 & 45.9 & 69.5 & 66.7 & 33.5 & 6.7 & 12.3 & 42.7 & 2.2 & 28.0 & 5.8 & 21.6 & 21.1 & 3.9 & 14.2 & 51.9 & 10.6 & 33.8 & 31.0 & 72.1 & 13.8 \\
                        MTML~\cite{lahoud20193d} & 28.2 & 54.9 & 73.1 & 57.7 & 38.0 & 18.2 & 10.7 & 43.0 & 0.1 & 42.2 & 5.7 & 17.9 & 16.2 & 7.0 & 22.9 & 51.1 & 16.1 & 49.1 & 31.3 & 65.0 & 16.2\\
                        3D-MPA~\cite{engelmann20203d} & 35.5 & 67.2 & 74.2 & 48.4 & 29.9 & 27.7 & 59.1 & 4.7 & \textbf{33.2} & 21.2 & 21.7 & \textbf{27.8} & 19.3 & \textbf{41.3} & 41.0 & 19.5 & \textbf{57.4} & 35.2 & 35.2 & 84.9 & 21.3 \\
                        SSEN~\cite{zhang2020ssen} & 38.4 & 57.5 & 72.4 & 85.2 & 49.4 & 19.2 & 22.6 & 64.8 & 2.2 & 39.8 & \textbf{29.9} & 27.7 & 31.7 & 23.1 & 19.4 & 51.4 & 19.6 & 58.6 & \textbf{44.4} & 84.3 & 18.4 \\
                        PE~\cite{zhang2021point} & 39.6 & \textbf{64.5} & 77.6 & \textbf{66.7} & 46.7 & 44.6 & 24.3 & 62.4 & 2.2 & 57.7 & 10.6 & 21.9 & 34.0 & 23.9 & \textbf{48.7} & 47.5 & 22.5 & 54.1 & 35.0 & 81.8 & 27.3 \\ 
                        PointGroup~\cite{pointgroup} & 40.7 & 63.6 & \textbf{77.8} & 63.9 & 49.6 & 41.5 & 24.3 & 64.5 & 2.1 & 57.0 & 11.4 & 21.1 & 35.9 & 21.7 & 42.8 & 66.0 & 25.6 & 56.2 & 34.1 & \textbf{86.0} & 29.1 \\
                        DyCo3D~\cite{dyco3d} & 39.5 & 64.1 & 76.1 & 64.2 & \textbf{51.8} & 44.7 & 25.9 & \textbf{66.6} & 5.0 & 25.1 & 16.6 & 23.1 & 36.2 & 23.2 & 33.1 & 53.5 & 22.9 & \textbf{58.7} & 43.8 & 85.0 & \textbf{31.7} \\
                        \textbf{Ours (DD-UNet+Group)} & \textbf{43.6} & 63.5 & 76.4 & 63.0 & 50.8 & \textbf{48.0} & \textbf{31.0} & 62.4 & 6.5 & \textbf{63.8} & 17.4 & 25.6 & \textbf{38.4} & 19.4 & 42.8 & \textbf{75.9} & 28.9 & 57.4 & 40.0 & 84.9 & 29.1\\
                        \bottomrule
                \end{tabular}
        }
        \vskip-2ex
        \caption{Per class mAP 3D instance segmentation results on ScanNet v2~\cite{scannet} testing set. mAP, $AP_{50}$ and $AP_{25}$ are reported.}
        \label{tab:scannet_test}
\vskip-3ex
\end{table*}}

Following previous researches, mean Average Precision (mAP) is leveraged in our work as the main evaluation metric (see Table~\ref{tab:scannet_test} and Table~\ref{tab:scannet_val}). Specifically, $AP_{25}$ and $AP_{50}$ denote the AP scores with IoU threshold set to $25\%$ and $50\%$. Also, AP averages the scores with IoU threshold set from $50\%$ to $95\%$, with a step size of $5\%$. We first assess the performance on the validation set of ScanNet v2, as shown in Table~\ref{tab:scannet_val}. Here, we also compare the performance with different backbone sizes: ``*-S'' denotes a smaller backbone with $m=16$ and ``*-L'' denotes a large backbone with $m=32$. If we focus on the smaller-size backbone and preform a fair comparison with the original PointGroup model~\cite{pointgroup} and the recent DyCo3D model~\cite{dyco3d}, the proposed architecture clearly exceeds them, \ie, by $2.8\%$ compared to PointGroup and $2.6\%$ compared to DyCo3D. 

We also report the class-wise performance of our 3D point cloud instance segmentation model on the testing set of ScanNet v2, as listed in Table~\ref{tab:scannet_test}.  Compared with state-of-the-art methods, our DD-UNet+Group model has achieved the best performance measured in mAP ($43.6\%$). Among the $11$ architectures, our method reaches high scores on many classes relevant for assisting the visually impaired.

\begin{table}[t]
    \setlength{\tabcolsep}{5mm}{
    \footnotesize
    \begin{center}
    \begin{tabular}{ r |c c c}
        \toprule
        &\textbf{mAP} &\textbf{$AP_{50}$} & \textbf{$AP_{25}$}\\
        \midrule
        MTML~\cite{lahoud20193d}& 20.3 & 40.2 & 55.4\\
        3D-MPA~\cite{engelmann20203d}& 35.3  & 59.1 & 72.4\\
        PointGroup~\cite{pointgroup}& 35.2& 57.1 & 71.4\\
        DyCo3D-S~\cite{dyco3d}& 35.4 & 57.6 & 72.9\\
        DyCo3D-L~\cite{dyco3d}& 40.6 & \textbf{61.0} & -\\
        \textbf{Ours-S} & 38.0 & 58.5 & 72.5\\
        \textbf{Ours-L} & \textbf{42.4} & 60.3 & \textbf{74.0}\\
        \bottomrule
    \end{tabular}
    \vskip-2ex
    \caption{Results on ScanNet v2~\cite{scannet} validation set.}
    \label{tab:scannet_val}
    \end{center}
    }
\vskip-4ex
\end{table}

\subsubsection{Runtime analysis of point cloud segmentation}
We tested this system on a laptop (Intel Core i7-7700HQ, NVIDIA GTX 1050Ti). We scanned four real-world scenes including a densely-scanned meeting room, a bedroom, a corridor, and an office. The scan time and point cloud sizes are different. We counted the pre-processing time, instance segmentation time (specific to each part of the network), and the extraction time of object information, listed in the Table~\ref{tab:runtime}. It can be seen that the processing of the point cloud is still computationally complex, which inevitably leads to a short waiting time for the user. In the scenario of helping the visually impaired gather complete scene understanding, search for interested objects, or travel in unfamiliar indoor scenes, a short waiting time is understandable. Yet, a more powerful mobile processor and a more efficient large-scale point cloud instance segmentation method are expected to allow users travelling instantly after the scanning. 

\begin{table}[t]
    \footnotesize
        \begin{center}
        \scalebox{0.8}[0.8]{
            \begin{tabular}{ c | c | c|| c | c | c | c | c | c}
                \toprule
                  & {\#Points} &  {\#Points*} & {Total} &  {PRE} &  {BB}  & {CL}  &  {S\&P}  &  {DET}\\
                \midrule
                1 & 819,073 & 163,806 & 9.515 & 2.661 & 1.709 & 0.179 & 0.502 & 3.929\\
                2 & 269,273  & 134,264 & 8.337 & 3.117 & 1.349 & 0.554 & 0.112 & 2.959\\
                3 & 169,791 & 169,364 & 9.785 & 2.551 & 1.504 & 0.250 & 0.556 & 4.429\\
                4 & 258,496 & 128,897 & 6.512 & 1.604 & 1.417 & 0.174 & 0.314 & 3.604\\
                \midrule
                avg & 379,158 & 149,083 & 8.788 & 2.483 & 1.495 & 0.289 & 0.371 & 3.729\\
                \bottomrule
            \end{tabular}
        }
        \end{center}
        \vskip-3ex
        \caption{Analysis of inference time (s).
                * denotes pre-processed point cloud;
                Total denotes the time from scan finished to detection done;
                PRE denotes outliers removing and downsampling;
                BB denotes backbone $+$ two branches;
                CL denotes the clustering part;
                S\&P denotes ScoreNet and final instance prediction;
                DET denotes information extraction from instances map.}
        \label{tab:runtime}
\vskip-3ex
\end{table}

\subsection{User Studies}

\subsubsection{Function test}
Two tasks are designed to test HIDA with $7$ participants including $5$ males and $2$ females. Their ages are within 24-30. Due to COVID-19 restrictions especially the social-distancing regulation challenging for the visually impaired~\cite{martinez2020helping,shrestha2020active}, the voluntary participants were sighted, and they were blindfolded during the tests. The first task focuses on the functionality of obstacle avoidance, whereas the second tests the functionality of instance locating and guidance ability for assisting users to find their interested objects. When each participant was executing their tasks, we recorded the whole process for subsequent analysis.

\noindent
\textbf{Task 1, find passable direction:}
The scenario of the first task is: The user enters a wide corridor (in most directions passable), within three meters around the entrance, we place different kinds of obstacles and leave a ``gap" in a random direction. The users were asked to find and leave this area with obstacles. Each user performed this task first using only the white cane and then performed using both the white cane and our system. The arrangement and position of obstacles and the gap will be changed after each task has been successfully executed.

\begin{figure*}[t]
\begin{center}
   \includegraphics[width=1\linewidth]{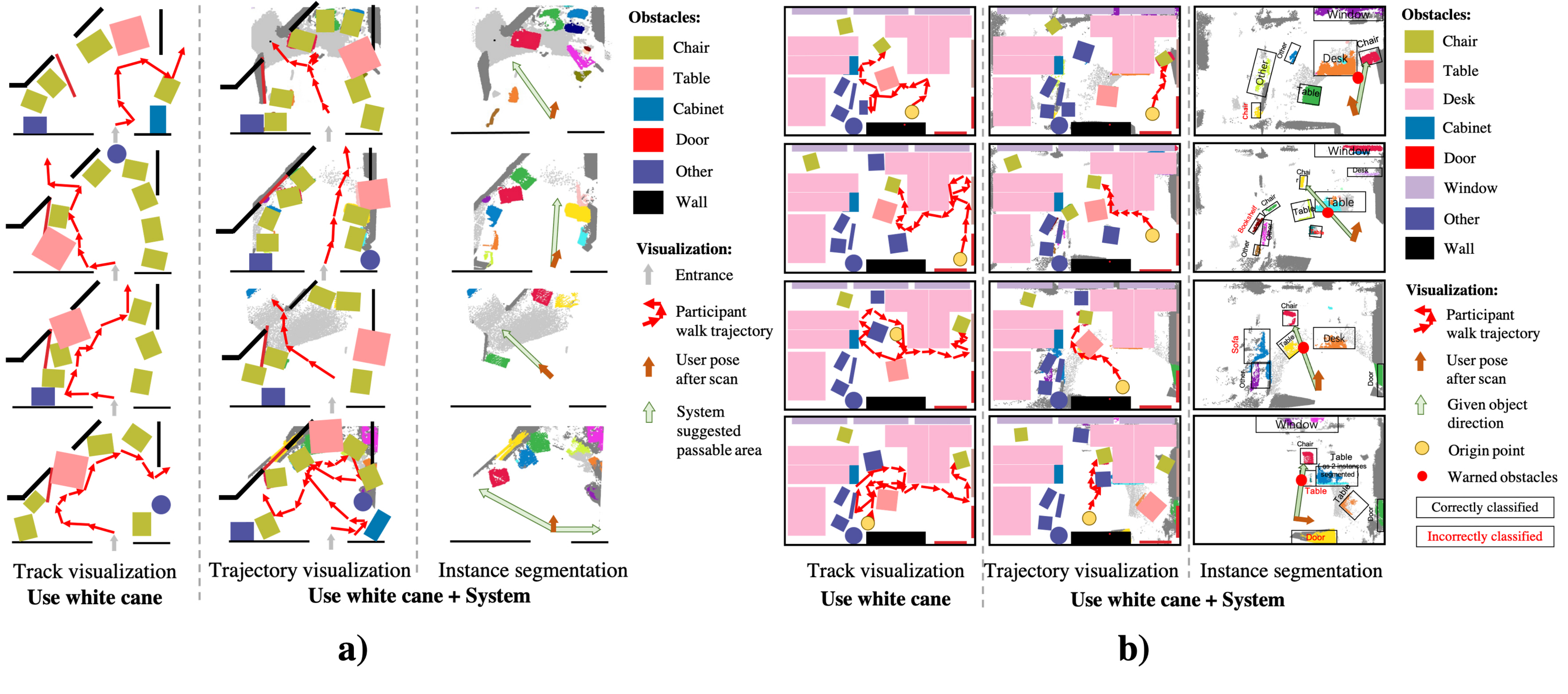}
\end{center}
   \vskip-6ex
   \caption{\textbf{a)}: Visualization of task 1; \textbf{b)}: Visualization of task 2. For each visualization, floor maps are plotted according to recorded videos and saved point clouds, and trajectories are manually plotted according to the recorded videos. Instance segmentation and the system output visualization are on the right.}
\label{task}
\vskip-3ex
\end{figure*}

Fig.~\ref{task}a) illustrates some of our qualitative analysis results of task 1. It can be seen that in the case of using only a white cane, users generally do global search without guidance, trying to recognize the surrounding environment and find passable areas by a brute-force like approach. While using HIDA, participants can directly find the correct direction of the gap. The collisions with obstacles were significantly reduced, providing a safe walking condition in indoor environments with obstacles. However, few problems among different users are found. For example, the vision-impaired user sometimes can not completely scan the surrounding environment. If our system did not find a ``gap'' between the scanned obstacles, it will suggest two directions pointed to unscanned area. The last row of Fig.~\ref{task}a) represents this case. Although there is a correct direction among two suggested directions, it is hard for users to go back to the previous position after they primarily tried in the wrong direction. This indicates that one should have more practice in order to maximize the system's effect.

\noindent
\textbf{Task 2, find a specific object:}
We chose an office for the second task since objects placement in this office seems more complex, indicating a kind of difficulty level for testing our system. First, users will be taken to different start points. Then, the users were asked to find a random \emph{chair} and sit down. Each user fulfilled this task twice, the first time with the white cane, and the second time with the white cane and our system. We only changed the position of part objects each time. In order to avoid the user becoming more familiar with the room when entering the room for the second time, half of the users first performed the task with white cane and then used both the system and cane, the other half on the contrary. 

Fig.~\ref{task}b) visualizes part of the results. Similar to task 1, when the user only relies on the white cane, the user perceived the surrounding objects one by one through touching them and feeling their shape, showing a low efficiency for finding the object. In the case of using our system as supplement, the users walked directly toward the chair, discarding the process of object shape recognition by users through touching, indicating a higher efficiency in searching objects. And they could bypass the obstacles on the path. As shown in the last row in Fig.~\ref{task}b), some objects may not be correctly classified since the accuracy of our segmentation network is not perfect. A network with an even higher precision is expected to be designed and deployed in the future.

\noindent\textbf{Efficiency analysis.}
We also counted the average time the user costed to complete these tasks in Table~\ref{tab:efficiency}. As it can be seen from the data, using our system did not significantly reduce the whole duration time. The reason is that scanning of the surrounding environment, segmentation of the point cloud, and audio interaction with users are still time-consuming for the current method. If we only measure the time from the user leaving the starting area to completing the task, the differences are clear, indicating the potential and superiority of the proposed system, which gave a correct direction and significantly reduced wrong attempts.

\begin{table}[b]
\vskip-1ex
	\centering
	    \small
		\begin{tabular}{c|c c c}
			\toprule[1pt]
			Task & Cane only & Cane+System & Cane+System *\\
			\hline
			1 & 48.3s & 61.5s & 21.5s\\
			2 & 70.2s & 72.6s & 26.0s\\
			\bottomrule[1pt]
		\end{tabular}
	\vskip-1ex
	\caption{Comparison of the time required to complete the task. * denotes ignoring the processing time.}
	\label{tab:efficiency}
\end{table}

\subsubsection{Feedback}
\begin{figure}[h]
    \begin{center}
    \scalebox{0.9}[0.9]{
        \includegraphics[width=1\linewidth]{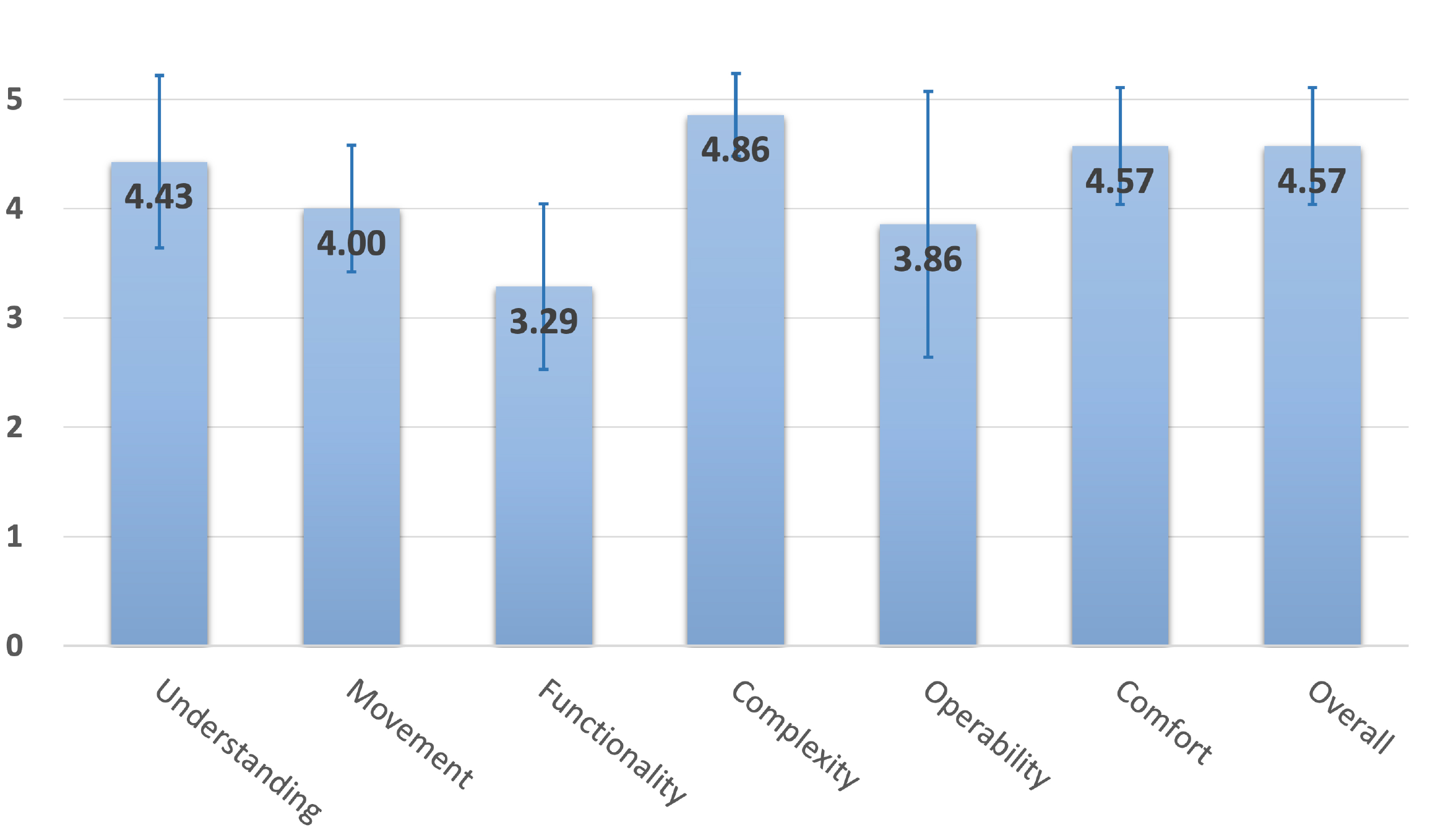}
    }\end{center}
    \caption{Questionnaire feedback from the participants, error bars indicate standard error.}
    \label{questionnaires}
\end{figure}

\noindent\textbf{Questionnaire.}
A questionnaire from multiple aspects, listed in Table~\ref{tab:questionnaire}, is designed in order to further evaluate the user experience of our system. After the participants completed the above two tasks, each participant answered these questions. Fig.~\ref{questionnaires} shows the feedback from the participants regarding our system. Participants make positive comments in terms of understanding, complexity, and comfort of the functionality of our system. They consider, that the system has an intuitive and smooth interface, and it is much easier to find the object with this system. However, the functionality and the operability still need to be improved. In summary, most participants think that the functions of HIDA are indeed very helpful in case of visual impairment and wearing the current hardware will not have too much impact on their daily lives. But they expect more functions, in addition, they think that some training are necessary in order to become familiar with the system, then they will accomplish the tasks more smoothly.

\begin{table}[t]
    \renewcommand\arraystretch{1.3}
	\centering
	    \footnotesize 
		\begin{tabular}{l|p{5cm}}
			\toprule[1pt]
			  & \textbf{Questionnaires}\\
			\hline
			Understanding & I can understand an indoor environment by using this system.\\
            Movement & I know where the obstacle located.\\
            Functionality & The system meets my expectations of what a navigation system can do.\\
            Complexity & I do not have to concentrate all my attention on the subtle changes of sound.\\
            Operability & I can get familiar with this system easily; The audio guidance is clear.\\
            Comfort & Wearing this device has no negative effect on my other daily actions.\\
            Overall & For vision impaired people, this system helps.\\
			\bottomrule[1pt]
		\end{tabular}
	\vskip-1ex
	\caption{Questionnaires: For each statement, users will select a score among 1-5, 5 means strongly agree, and 1 means strongly disagree.}
	\label{tab:questionnaire}
\end{table}

\noindent\textbf{User comments.}
We also collected some comments from the participants. Some positive comments include that our system did help them to understand the environment, it provided a very smooth interface, it is easy to wear, and the bone-conduction headset is also very suitable for such a system. However, some participants still put forward some improvement directions of this system. In the hardware aspect, one participant hopes to use a lighter processor instead of laptop in the future, which will make our system more wearable. In the functional aspect, two participants both suggested that if the real-time objects information could be provided, it will help the visually impaired more. In the interaction aspect, one participant mentioned that the current voice interaction is still a bit time-consuming, especially when there are many obstacles, the output of information is somewhat lengthy. In addition, some participants thought this system should have a clearer guidance on how to scan a room completely or give more training before the usage. These comments are very constructive for our future work.

\section{Conclusion and Future Works}

In this work, HIDA, a 3D vision-based assistance system is proposed to help visually impaired people gather a holistic indoor understanding with object detection and obstacle avoidance. By using point cloud obtained with a wearable solid-state LiDAR sensor, it provides obstacle detection and specific object searching. Visual SLAM and instance-aware 3D scene parsing are combined, where point-wise semantics are converted to yield a dense top-view surrounding perception. The devised 3D instance segmentation model attains state-of-the-art performance on ScanNet v2 dataset, thanks to separate predictions of semantic- and offset features. The overall system is verified to be useful and helpful for indoor travelling and object searching, and it is promising for more assisted living tasks.

However, there are some limitations of our system. First, the surrounding environment scanning and point cloud segmentation are time-consuming, making a real-time processing not possible. In addition, the accuracy of the current point cloud instance segmentation model still has much space to be improved. Last but not least, although we have obtained a satisfactory point cloud map with semantic instance information, due to the visual odometry get lost easily when the movement range is too large, the real-time interaction with the map has not been fully investigated.
 
In the future, we intend to address the above points and introduce improvements to the system such as leveraging multi-sensor fusion to speed up the scanning of the surrounding environment and improving the odometry to obtain the location of the user on the map in real time. Moreover, designing a higher-precision segmentation network structure will also significantly improve the accuracy of holistic guidance delivered by our system.

\clearpage

{\small
\bibliographystyle{ieee_fullname}
\bibliography{egbib}

\begin{thebibliography}{10}\itemsep=-1pt

\bibitem{aladren2014navigation}
Aitor Aladren, Gonzalo L{\'{o}}pez{-}Nicol{\'{a}}s, Luis Puig, and Josechu~J.
  Guerrero.
\newblock Navigation assistance for the visually impaired using {RGB-D} sensor
  with range expansion.
\newblock {\em IEEE Systems Journal}, 2016.

\bibitem{S3DIS}
Iro Armeni, Ozan Sener, Amir~Roshan Zamir, Helen Jiang, Ioannis~K. Brilakis,
  Martin Fischer, and Silvio Savarese.
\newblock {3D} semantic parsing of large-scale indoor spaces.
\newblock In {\em CVPR}, 2016.

\bibitem{bai2017smart}
Jinqiang Bai, Shiguo Lian, Zhaoxiang Liu, Kai Wang, and Dijun Liu.
\newblock Smart guiding glasses for visually impaired people in indoor
  environment.
\newblock {\em IEEE Transactions on Consumer Electronics}, 2017.

\bibitem{bauer2020enhancing}
Zuria Bauer, Alejandro Dominguez, Edmanuel Cruz, Francisco Gomez-Donoso, Sergio
  Orts-Escolano, and Miguel Cazorla.
\newblock Enhancing perception for the visually impaired with deep learning
  techniques and low-cost wearable sensors.
\newblock {\em Pattern Recognition Letters}, 2020.

\bibitem{cao2020rapid}
Zhengcai Cao, Xiaowen Xu, Biao Hu, and MengChu Zhou.
\newblock Rapid detection of blind roads and crosswalks by using a lightweight
  semantic segmentation network.
\newblock {\em IEEE Transactions on Intelligent Transportation Systems}, 2020.

\bibitem{caraiman2017computer}
Simona Caraiman, Anca Morar, Mateusz Owczarek, Adrian Burlacu, Dariusz
  Rzeszotarski, Nicolae Botezatu, Paul Herghelegiu, Florica Moldoveanu, Pawel
  Strumillo, and Alin Moldoveanu.
\newblock Computer vision for the visually impaired: the sound of vision
  system.
\newblock In {\em ICCVW}, 2017.

\bibitem{cartillier2020semantic}
Vincent Cartillier, Zhile Ren, Neha Jain, Stefan Lee, Irfan Essa, and Dhruv
  Batra.
\newblock Semantic {MapNet}: {Building} allocentric {SemanticMaps} and
  representations from egocentric views.
\newblock In {\em AAAI}, 2021.

\bibitem{chen2021panoramic}
Hao Chen, Weijian Hu, Kailun Yang, Jian Bai, and Kaiwei Wang.
\newblock Panoramic annular {SLAM} with loop closure and global optimization.
\newblock {\em Applied Optics}, 2021.

\bibitem{scannet}
Angela Dai, Angel~X. Chang, Manolis Savva, Maciej Halber, Thomas~A. Funkhouser,
  and Matthias Nie{\ss}ner.
\newblock {ScanNet:} richly-annotated {3D} reconstructions of indoor scenes.

\bibitem{duh2020v}
Ping-Jung Duh, Yu-Cheng Sung, Liang-Yu~Fan Chiang, Yung-Ju Chang, and Kuan-Wen
  Chen.
\newblock {V-Eye:} {A} vision-based navigation system for the visually
  impaired.
\newblock {\em IEEE Transactions on Multimedia}, 2020.

\bibitem{engelmann20203d}
Francis Engelmann, Martin Bokeloh, Alireza Fathi, Bastian Leibe, and Matthias
  Nie{\ss}ner.
\newblock {3D-MPA:} {Multi-proposal} aggregation for {3D} semantic instance
  segmentation.
\newblock In {\em CVPR}, 2020.

\bibitem{omni_point}
Jingyu Gong, Jiachen Xu, Xin Tan, Haichuan Song, Yanyun Qu, Yuan Xie, and
  Lizhuang Ma.
\newblock Omni-supervised point cloud segmentation via gradual receptive field
  component reasoning.
\newblock In {\em CVPR}, 2021.

\bibitem{guo2020pct}
Meng{-}Hao Guo, Junxiong Cai, Zheng{-}Ning Liu, Tai{-}Jiang Mu, Ralph~R.
  Martin, and Shi{-}Min Hu.
\newblock {PCT:} {Point} cloud transformer.
\newblock {\em Computational Visual Media}, 2021.

\bibitem{han2020occuseg}
Lei Han, Tian Zheng, Lan Xu, and Lu Fang.
\newblock {OccuSeg:} {Occupancy-aware} {3D} instance segmentation.
\newblock In {\em CVPR}, 2020.

\bibitem{he2017mask}
Kaiming He, Georgia Gkioxari, Piotr Doll{\'a}r, and Ross Girshick.
\newblock Mask {R-CNN}.
\newblock In {\em ICCV}, 2017.

\bibitem{dyco3d}
Tong He, Chunhua Shen, and Anton van~den Hengel.
\newblock {DyCo3D:} {Robust} instance segmentation of {3D} point clouds through
  dynamic convolution.
\newblock In {\em CVPR}, 2021.

\bibitem{hou20193d}
Ji Hou, Angela Dai, and Matthias Nie{\ss}ner.
\newblock {3D-SIS:} {3D} semantic instance segmentation of {RGB-D} scans.
\newblock In {\em CVPR}, 2019.

\bibitem{hsieh2020outdoor}
I-Hsuan Hsieh, Hsiao-Chu Cheng, Hao-Hsiang Ke, Hsiang-Chieh Chen, and Wen-June
  Wang.
\newblock Outdoor walking guide for the visually-impaired people based on
  semantic segmentation and depth map.
\newblock In {\em ICPAI}, 2020.

\bibitem{hsieh2020development}
Yi-Zeng Hsieh, Shih-Syun Lin, and Fu-Xiong Xu.
\newblock Development of a wearable guide device based on convolutional neural
  network for blind or visually impaired persons.
\newblock {\em Multimedia Tools and Applications}, 2020.

\bibitem{hu2019indoor}
Weijian Hu, Kaiwei Wang, Hao Chen, Ruiqi Cheng, and Kailun Yang.
\newblock An indoor positioning framework based on panoramic visual odometry
  for visually impaired people.
\newblock {\em Measurement Science and Technology}, 2019.

\bibitem{hu2021bidirectional}
Wenbo Hu, Hengshuang Zhao, Li Jiang, Jiaya Jia, and Tien-Tsin Wong.
\newblock Bidirectional projection network for cross dimension scene
  understanding.
\newblock In {\em CVPR}, 2021.

\bibitem{jaus2021panoramic}
Alexander Jaus, Kailun Yang, and Rainer Stiefelhagen.
\newblock Panoramic panoptic segmentation: Towards complete surrounding
  understanding via unsupervised contrastive learning.
\newblock In {\em IV}, 2021.

\bibitem{pointgroup}
Li Jiang, Hengshuang Zhao, Shaoshuai Shi, Shu Liu, Chi-Wing Fu, and Jiaya Jia.
\newblock {PointGroup:} {Dual-set} point grouping for {3D} instance
  segmentation.
\newblock In {\em CVPR}, 2020.

\bibitem{katzschmann2018safe}
Robert~K. Katzschmann, Brandon Araki, and Daniela Rus.
\newblock Safe local navigation for visually impaired users with a
  time-of-flight and haptic feedback device.
\newblock {\em IEEE Transactions on Neural Systems and Rehabilitation
  Engineering}, 2018.

\bibitem{adam_optimization}
Diederik~P. Kingma and Jimmy Ba.
\newblock Adam: A method for stochastic optimization.
\newblock In {\em ICLR}, 2015.

\bibitem{labbe2019rtab}
Mathieu Labb{\'e} and Fran{\c{c}}ois Michaud.
\newblock {RTAB-Map} as an open-source lidar and visual simultaneous
  localization and mapping library for large-scale and long-term online
  operation.
\newblock {\em Journal of Field Robotics}, 2019.

\bibitem{lahoud20193d}
Jean Lahoud, Bernard Ghanem, Martin~R. Oswald, and Marc Pollefeys.
\newblock {3D} instance segmentation via multi-task metric learning.
\newblock In {\em ICCV}, 2019.

\bibitem{li2016isana}
Bing Li, J.~Pablo Mu{\~{n}}oz, Xuejian Rong, Jizhong Xiao, Yingli Tian, and
  Aries Arditi.
\newblock {ISANA:} {Wearable} context-aware indoor assistive navigation with
  obstacle avoidance for the blind.
\newblock In {\em ECCVW}, 2016.

\bibitem{li2018semantic}
Ruihao Li, Dongbing Gu, Qiang Liu, Zhiqiang Long, and Huosheng Hu.
\newblock Semantic scene mapping with spatio-temporal deep neural network for
  robotic applications.
\newblock {\em Cognitive Computation}, 2018.

\bibitem{liang20203d}
Zhidong Liang, Ming Yang, Hao Li, and Chunxiang Wang.
\newblock {3D} instance embedding learning with a structure-aware loss function
  for point cloud segmentation.
\newblock {\em IEEE Robotics and Automation Letters}, 2020.

\bibitem{lin2019deep}
Yimin Lin, Kai Wang, Wanxin Yi, and Shiguo Lian.
\newblock Deep learning based wearable assistive system for visually impaired
  people.
\newblock In {\em ICCVW}, 2019.

\bibitem{liu2019masc}
Chen Liu and Yasutaka Furukawa.
\newblock {MASC:} {Multi-scale} affinity with sparse convolution for {3D}
  instance segmentation.
\newblock {\em arXiv}, 2019.

\bibitem{liu2020indoor}
Qiang Liu, Ruihao Li, Huosheng Hu, and Dongbing Gu.
\newblock Indoor topological localization based on a novel deep learning
  technique.
\newblock {\em Cognitive Computation}, 2020.

\bibitem{liu20213d}
Zhengzhe Liu, Xiaojuan Qi, and Chi-Wing Fu.
\newblock {3D-to-2D} distillation for indoor scene parsing.
\newblock In {\em CVPR}, 2021.

\bibitem{fcn}
Jonathan Long, Evan Shelhamer, and Trevor Darrell.
\newblock Fully convolutional networks for semantic segmentation.
\newblock In {\em CVPR}, 2015.

\bibitem{long2019unifying}
Ningbo Long, Kaiwei Wang, Ruiqi Cheng, Weijian Hu, and Kailun Yang.
\newblock Unifying obstacle detection, recognition, and fusion based on
  millimeter wave radar and {RGB-depth} sensors for the visually impaired.
\newblock {\em Review of Scientific Instruments}, 2019.

\bibitem{mahendran2021computer}
Jagadish~K. Mahendran, Daniel~T. Barry, Anita~K. Nivedha, and Suchendra~M.
  Bhandarkar.
\newblock Computer vision-based assistance system for the visually impaired
  using mobile edge artificial intelligence.
\newblock In {\em CVPRW}, 2021.

\bibitem{mao2021panoptic}
Wei Mao, Jiaming Zhang, Kailun Yang, and Rainer Stiefelhagen.
\newblock Panoptic lintention network: Towards efficient navigational
  perception for the visually impaired.
\newblock In {\em RCAR}, 2021.

\bibitem{martinez2020helping}
Manuel Martinez, Kailun Yang, Angela Constantinescu, and Rainer Stiefelhagen.
\newblock Helping the blind to get through {COVID-19:} {Social} distancing
  assistant using real-time semantic segmentation on {RGB-D} video.
\newblock {\em Sensors}, 2020.

\bibitem{miksik2015semantic}
Ondrej Miksik, Vibhav Vineet, Morten Lidegaard, Ram Prasaath, Matthias
  Nie{\ss}ner, Stuart Golodetz, Stephen~L. Hicks, Patrick P{\'{e}}rez, Shahram
  Izadi, and Philip H.~S. Torr.
\newblock The semantic paintbrush: Interactive {3D} mapping and recognition in
  large outdoor spaces.
\newblock In {\em CHI}, 2015.

\bibitem{narita2019panopticfusion}
Gaku Narita, Takashi Seno, Tomoya Ishikawa, and Yohsuke Kaji.
\newblock {PanopticFusion:} {Online} volumetric semantic mapping at the level
  of stuff and things.
\newblock In {\em IROS}, 2019.

\bibitem{pan2020cross}
Bowen Pan, Jiankai Sun, Ho~Yin~Tiga Leung, Alex Andonian, and Bolei Zhou.
\newblock Cross-view semantic segmentation for sensing surroundings.
\newblock {\em IEEE Robotics and Automation Letters}, 2020.

\bibitem{peng2021mass}
Kunyu Peng, Juncong Fei, Kailun Yang, Alina Roitberg, Jiaming Zhang, Frank
  Bieder, Philipp Heidenreich, Christoph Stiller, and Rainer Stiefelhagen.
\newblock {MASS:} {Multi-attentional} semantic segmentation of {LiDAR} data for
  dense top-view understanding.
\newblock {\em arXiv}, 2021.

\bibitem{perez2017stairs}
Alejandro P{\'{e}}rez{-}Yus, Daniel Guti{\'{e}}rrez{-}G{\'{o}}mez, Gonzalo
  L{\'{o}}pez{-}Nicol{\'{a}}s, and J.~J. Guerrero.
\newblock Stairs detection with odometry-aided traversal from a wearable
  {RGB-D} camera.
\newblock {\em Computer Vision and Image Understanding}, 2017.

\bibitem{pham2019jsis3d}
Quang-Hieu Pham, Thanh Nguyen, Binh-Son Hua, Gemma Roig, and Sai-Kit Yeung.
\newblock {JSIS3D:} {Joint} semantic-instance segmentation of {3D} point clouds
  with multi-task pointwise networks and multi-value conditional random fields.
\newblock In {\em CVPR}, 2019.

\bibitem{ronneberger2015u}
Olaf Ronneberger, Philipp Fischer, and Thomas Brox.
\newblock {U-Net:} {Convolutional} networks for biomedical image segmentation.
\newblock In {\em MICCAI}, 2015.

\bibitem{shrestha2020active}
Samridha Shrestha, Daohan Lu, Hanlin Tian, Qiming Cao, Julie Liu, John-Ross
  Rizzo, William~H. Seiple, Maurizio Porfiri, and Yi Fang.
\newblock Active crowd analysis for pandemic risk mitigation for blind or
  visually impaired persons.
\newblock In {\em ECCVW}, 2020.

\bibitem{wang2017enabling}
Hsueh{-}Cheng Wang, Robert~K. Katzschmann, Santani Teng, Brandon Araki, Laura
  Giarr{\'{e}}, and Daniela Rus.
\newblock Enabling independent navigation for visually impaired people through
  a wearable vision-based feedback system.
\newblock In {\em ICRA}, 2017.

\bibitem{wang2021lightweight}
Han Wang, Chen Wang, and Lihua Xie.
\newblock Lightweight {3-D} localization and mapping for solid-state {LiDAR}.
\newblock {\em IEEE Robotics and Automation Letters}, 2021.

\bibitem{wang2018environmental}
Juan Wang, Kailun Yang, Weijian Hu, and Kaiwei Wang.
\newblock An environmental perception and navigational assistance system for
  visually impaired persons based on semantic stixels and sound interaction.
\newblock In {\em SMC}, 2018.

\bibitem{wang2018sgpn}
Weiyue Wang, Ronald Yu, Qiangui Huang, and Ulrich Neumann.
\newblock {SGPN:} {Similarity} group proposal network for {3D} point cloud
  instance segmentation.
\newblock In {\em CVPR}, 2018.

\bibitem{wang2019associatively}
Xinlong Wang, Shu Liu, Xiaoyong Shen, Chunhua Shen, and Jiaya Jia.
\newblock Associatively segmenting instances and semantics in point clouds.
\newblock In {\em CVPR}, 2019.

\bibitem{watson2020footprints}
Jamie Watson, Michael Firman, Aron Monszpart, and Gabriel~J Brostow.
\newblock Footprints and free space from a single color image.
\newblock In {\em CVPR}, 2020.

\bibitem{wu2021scenegraphfusion}
Shun-Cheng Wu, Johanna Wald, Keisuke Tateno, Nassir Navab, and Federico
  Tombari.
\newblock {SceneGraphFusion:} {Incremental} {3D} scene graph prediction from
  {RGB-D} sequences.
\newblock In {\em CVPR}, 2021.

\bibitem{yang2019learning}
Bo Yang, Jianan Wang, Ronald Clark, Qingyong Hu, Sen Wang, Andrew Markham, and
  Niki Trigoni.
\newblock Learning object bounding boxes for {3D} instance segmentation on
  point clouds.
\newblock In {\em NeurIPS}, 2019.

\bibitem{yang2018unifying}
Kailun Yang, Luis~Miguel Bergasa, Eduardo Romera, Ruiqi Cheng, Tianxue Chen,
  and Kaiwei Wang.
\newblock Unifying terrain awareness through real-time semantic segmentation.
\newblock In {\em IV}, 2018.

\bibitem{yang2020omnisupervised}
Kailun Yang, Xinxin Hu, Yicheng Fang, Kaiwei Wang, and Rainer Stiefelhagen.
\newblock Omnisupervised omnidirectional semantic segmentation.
\newblock {\em IEEE Transactions on Intelligent Transportation Systems}, 2020.

\bibitem{yang2015new}
Kailun Yang, Kaiwei Wang, Ruiqi Cheng, and Xunmin Zhu.
\newblock A new approach of point cloud processing and scene segmentation for
  guiding the visually impaired.
\newblock In {\em ICBISP}, 2015.

\bibitem{ye20173}
Cang Ye and Xiangfei Qian.
\newblock {3-D} object recognition of a robotic navigation aid for the visually
  impaired.
\newblock {\em IEEE Transactions on Neural Systems and Rehabilitation
  Engineering}, 2018.

\bibitem{yi2019gspn}
Li Yi, Wang Zhao, He Wang, Minhyuk Sung, and Leonidas~J Guibas.
\newblock {GSPN:} {Generative} shape proposal network for {3D} instance
  segmentation in point cloud.
\newblock In {\em CVPR}, 2019.

\bibitem{yohannes2019content}
Ervin Yohannes, Timothy~K. Shih, and Chih{-}Yang Lin.
\newblock Content-aware video analysis to guide visually impaired walking on
  the street.
\newblock In {\em IVIC}, 2019.

\bibitem{zatout2019ego}
Chayma Zatout, Slimane Larabi, Ilyes Mendili, and Soedji Ablam Edoh~Barnabe.
\newblock Ego-semantic labeling of scene from depth image for visually impaired
  and blind people.
\newblock In {\em ICCVW}, 2019.

\bibitem{zhang2021point}
Biao Zhang and Peter Wonka.
\newblock Point cloud instance segmentation using probabilistic embeddings.
\newblock In {\em CVPR}, 2021.

\bibitem{zhang2020ssen}
Dongsu Zhang, Junha Chun, Sang Cha, and Young~Min Kim.
\newblock Spatial semantic embedding network: Fast {3D} instance segmentation
  with deep metric learning.
\newblock In {\em arXiv}, 2020.

\bibitem{zhang2017indoor}
He Zhang and Cang Ye.
\newblock An indoor wayfinding system based on geometric features aided graph
  {SLAM} for the visually impaired.
\newblock {\em IEEE Transactions on Neural Systems and Rehabilitation
  Engineering}, 2017.

\bibitem{zhang2021perception}
Yingzhi Zhang, Haoye Chen, Kailun Yang, Jiaming Zhang, and Rainer Stiefelhagen.
\newblock Perception framework through real-time semantic segmentation and
  scene recognition on a wearable system for the visually impaired.
\newblock In {\em RCAR}, 2021.

\bibitem{zhao2020point}
Hengshuang Zhao, Li Jiang, Jiaya Jia, Philip H.~S. Torr, and Vladlen Koltun.
\newblock Point transformer.
\newblock {\em arXiv}, 2020.

\end{thebibliography}
}

\end{document}